\documentclass[11pt,journal,compsoc]{IEEEtran}

\usepackage{orcidlink}

\usepackage{silence}

\usepackage{titlesec}
\usepackage{amsmath,amssymb,amsfonts}
\usepackage{graphicx}
\usepackage{pifont}

\usepackage[nolist,nohyperlinks]{acronym}
\usepackage{color,soul}
\usepackage{amsthm}
\usepackage{xcolor}
\usepackage[figuresright]{rotating}

\usepackage{graphicx}
\graphicspath{{/}{fig/}}

\usepackage{array}
\usepackage{textcomp}
\usepackage{xcolor}
\usepackage{multirow}
\usepackage{booktabs}

\usepackage{mathtools}
\usepackage{breqn}
  \usepackage[frozencache=true,cachedir=minted-cache]{minted} 
\usepackage{pgfplots}
\pgfplotsset{compat=1.7}
\usepgfplotslibrary{groupplots}

\usepackage{caption}
\usepackage{subcaption}
\usepackage{enumerate}

\usepackage{multirow,tabularx}
\usepackage[vlined, ruled, shortend]{algorithm2e}
\usepackage{hyperref}
\usepackage{pbalance}

\newlength\figureheight
\newlength\figurewidth
\usepackage{cite}

\begin{document}

\title{Sim-to-Real Transfer for Mobile Robots with Reinforcement Learning: from NVIDIA Isaac Sim to Gazebo and Real ROS\,2 Robots}

\author{
  Sahar Salimpour, Jorge~Pe\~na-Queralta, Diego Paez-Granados, \\ Jukka Heikkonen, and Tomi Westerlund
  \IEEEcompsocitemizethanks{
    \IEEEcompsocthanksitem S. Salimpour, J. Heikkonen, and T. Westerlund were with the University of Turku, Finland. \protect\\
    Emails: \{sahars, jukhei, tovewe\}@utu.fi
    \IEEEcompsocthanksitem J. Pe\~na-Queralta and D. Paez-Granados were with the Swiss Federal School of Technology in Zurich - ETH Zurich, 8092 Switzerland. \protect\\
    Emails: \{jorge.penaqueralta, diego.paez\}@hest.ethz.ch.
  }
}

% \markboth{IEEE Robotics and Automation Magazine}%
% {TODO \MakeLowercase{\textit{et al.}}: ROS 2 HealthCare}

%%%%%%%%%%%%%%%%%%%%%%%%%%%%%%%%%%%%%%%%%%%%%%
%%                                          %%
%%          ABSTRACT, MAKE TITLE            %%
%%                                          %%
%%%%%%%%%%%%%%%%%%%%%%%%%%%%%%%%%%%%%%%%%%%%%%
\IEEEtitleabstractindextext{%
    
\begin{abstract}
    Unprecedented agility and dexterous manipulation have been demonstrated with controllers based on deep reinforcement learning (RL), with a significant impact on legged and humanoid robots. Modern tooling and simulation platforms, such as NVIDIA Isaac Sim, have been enabling such advances.  This article focuses on demonstrating the applications of Isaac in local planning and obstacle avoidance as one of the most fundamental ways in which a mobile robot interacts with its environments. Although there is extensive research on proprioception-based RL policies, the article highlights less standardized and reproducible approaches to exteroception. At the same time, the article aims to provide a base framework for end-to-end local navigation policies and how a custom robot can be trained in such simulation environment. We benchmark end-to-end policies with the state-of-the-art Nav2, navigation stack in Robot Operating System (ROS). We also cover the sim-to-real transfer process by demonstrating zero-shot transferability of policies trained in the Isaac simulator to real-world robots. This is further evidenced by the tests with different simulated robots, which show the generalization of the learned policy. Finally, the benchmarks demonstrate comparable performance to Nav2, opening the door to quick deployment of state-of-the-art end-to-end local planners for custom robot platforms, but importantly furthering the possibilities by expanding the state and action spaces or task definitions for more complex missions. Overall, with this article we introduce the most important steps, and aspects to consider, in deploying RL policies for local path planning and obstacle avoidance with Isaac Sim training, Gazebo testing, and ROS\,2 for real-time inference in real robots. The code is available at \url{https://github.com/sahars93/RL-Navigation}.
\end{abstract}

}

% make the title area
\maketitle

\begin{acronym}[] 

\acro{posa}[POSA]{Positional Obstructive Sleep Apnea}

\acro{ros2hc}[ROS2HC]{ROS to Healthcare}

\end{acronym}

\IEEEdisplaynontitleabstractindextext
\IEEEpeerreviewmaketitle

%%%%%%%%%%%%%%%%%%%%%%%%%%%%%%%%%%%%%%%%%%%%%%
%%                                          %%
%%                SECTIONS                  %%
%%                                          %%
%%%%%%%%%%%%%%%%%%%%%%%%%%%%%%%%%%%%%%%%%%%%%%

\IEEEraisesectionheading{\section{Introduction}\label{sec:introduction}}

% \noindent\red{\footnotesize 1) No more than 4500 words of text (~6 full pages)} \\
% \red{\footnotesize 2) No more than 10 equations} \\
% \red{\footnotesize 3) No more than 20 references} \\
% \red{\footnotesize 4) No more than 10 figures} \\

Reinforcement learning (RL) stands at the forefront of enabling complex control and facilitating advanced behaviors in various types of robots. This advancement holds promise for revolutionizing robotics, empowering machines to interact with their environments. RL algorithms are widely used across classic tasks including locomotion, navigation, or manipulation, among others. Indeed, recent years have seen unprecedented improvements in the ability of quadruped robots~\cite{gangapurwala2022rloc}, wheeled-legged robots~\cite{lee2024learning}, drone racing~\cite{song2023reaching}, humanoids~\cite{radosavovic2024real}, or bipedal robot sports~\cite{lee2024learning}. Also, in the automation of machinery such as hydraulic excavators~\cite{egli2020towards}.

In the majority of these problems, a policy is trained to map a control input, together with the robot sensory inputs, to join-level actuation. For example, \cite{gangapurwala2022rloc} maps from body-state velocities to steps, while \cite{radosavovic2024real} focuses on whole-body motion planning. Throughout this large variety of use cases and robotic systems, often focused on dexterous manipulation, or motion planning with a large number of degrees of freedom and/or uncertainty~\cite{han2023survey, margolis2024rapid, yang2022safe}, the field has established a range of commonly used approaches and, importantly, simulation tools. The latter include NVIDIA Isaac Sim or Orbit~\cite{mittal2023orbit, zhao2023less}, MuJoCo~\cite{radosavovic2024real, lee2024learning}, or Flightmare~\cite{song2023reaching}, among others. 

\begin{figure*}[t]
    \centering
    \includegraphics[width=\textwidth]{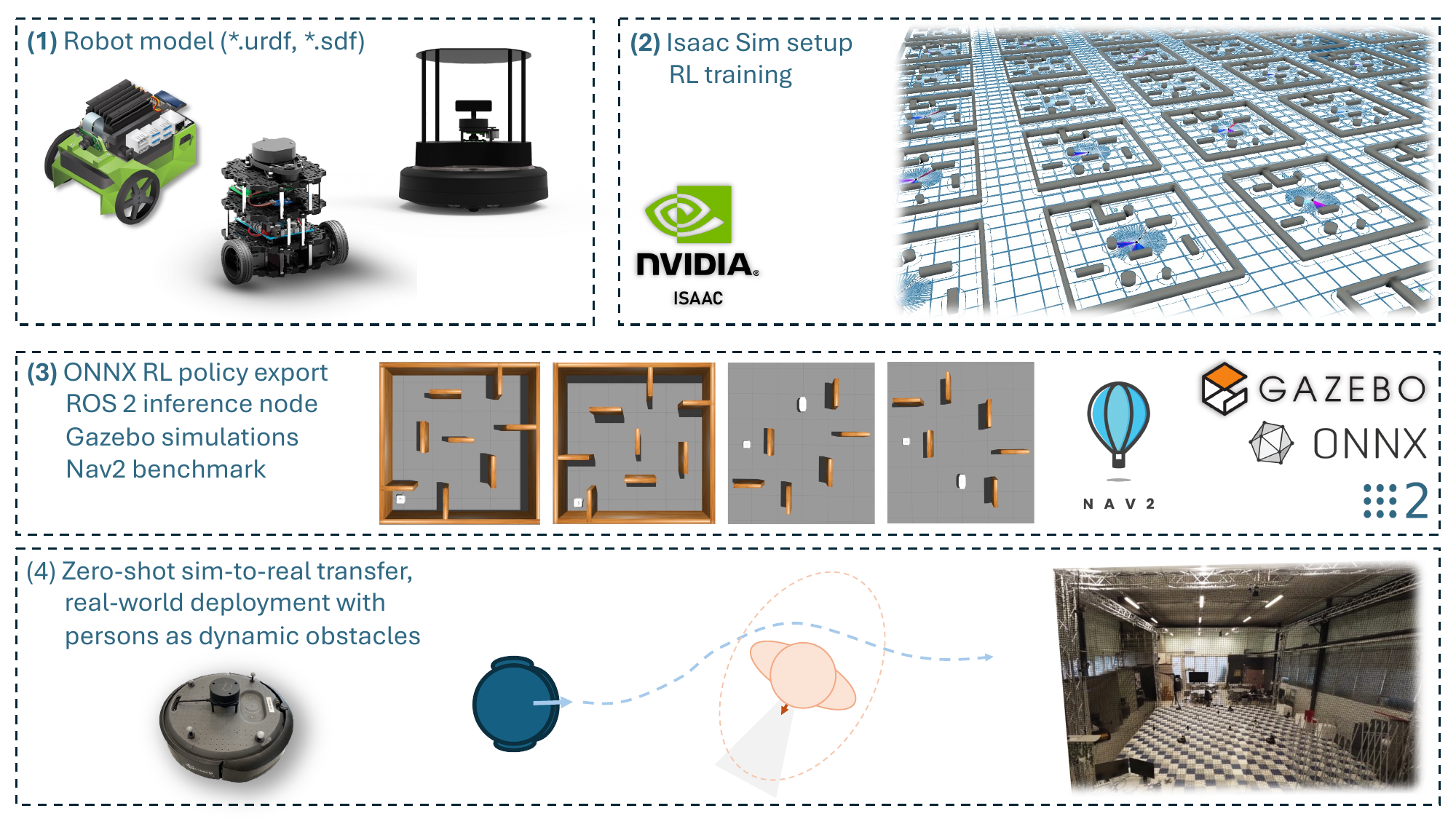}
    \caption{Conceptual illustration of the sim-to-real workflow described in this article. In the first step, we utilize different existing robot models, while also describing the Isaac model importer functionality in Section 3. In the second step, we describe key considerations in terms of RL policy training in Section 4, and the setup of different static and dynamic environments. In the third step, we provide template Robot Operating System (ROS\,2) nodes, and guidance on Gazebo testing in Sections 4 and 5. Additionally, we benchmark the performance to the state-of-the-art Nav2 navigation and planning algorithms. Finally, in the fourth step, we also demonstrate the zero-shot sim-to-real transfer capabilities in Section 5.}
    \label{fig:diagram}
\end{figure*}

Beyond low-level control and motion planning from propioceptive sensory inputs, RL has also been studied within the more general perspective of mobile robotics for local or global navigation, and high-level planning and autonomy. This applies to both complex robots such as quadrupeds~\cite{gangapurwala2022rloc}, to wheeled robots there motion planning is more straightforward. The level of standardization and the depth of the study of sim-to-real transfer for navigation task, however, is shallower~\cite{zhao2020sim}. Through this article, we aim to give more insight into such an a priori more rudimentary and classical problem, but where RL controllers can also play an important role as the field solidifies. Our focus is on providing an step-by-step approach to training RL policies for mobile robot navigation \textit{from scratch}, and describing the transition \textit{from simulation to reality} (see Figure~\ref{fig:diagram}).

% , including joint angles, velocities, and positions, are learned to map joint position commands for controllers in robotic manipulation, including grasping and object manipulation~\cite{han2023survey} and controlling legged robots~\cite{margolis2024rapid, yang2022safe}. In many of these cases, RL is used for motion control. However, for a general mobile robot, RL can also be applied to navigation and higher-level planning and autonomy, as motion control is relatively simple in wheeled and other typical locomotion solutions.

From this point forward, we constraint use cases to path planning and local navigation for mobile robots from an initial position to a target destination. The sensory focus is exterioceptive, aiming at training end-to-end policies that enable navigation without collisions in both simulated and real-world scenarios. This area, including complex static and dynamic environments, has been the subject of extensive study in recent literature~\cite{aradi2020survey}. 
% Traditional planning approaches require a model of the environment, such as map or the identification of specific obstacles for building.
% Conventional navigation methods like Simultaneous Localization and Mapping (SLAM) techniques often rely on static models of the environment or are tailored to specific types of robots~\cite{beomsoo2021mobile}. However, these approaches may not be sufficient in real-world scenarios where the environment is unknown and constantly changing. 
% This highlights the need for reliable, adaptable, and easy-to-implement methods that can navigate different types of mobile robots through unknown environments using local perception only. 
Deep reinforcement learning algorithms have emerged as a promising solution to this challenge~\cite{choi2021reinforcement}. Such RL algorithms have shown significant potential in navigation tasks for different types of robots and sensors such as LiDAR, RGB camera, and RGB-D camera~\cite{li2023navigation}. 

Some representative examples are the following.\cite{surmann2020deep}~proposes the usage of the Advantage Actor-Critic algorithm to navigate their robot in an environment with 3D obstacle avoidance, achieved through the fusion of a 2D laser scanner with an RGB-D camera in a self-implemented simulator. In ~\cite{choi2021reinforcement}, a soft actor-critic algorithm has been utilized to train and test an obstacle avoidance model for a differential drive robot. Similar to most lidar-based navigation studies, factors such as relative distance to the target point, lidar scan data, and the robot's speed are employed to determine the velocity necessary to drive toward the target point. Many of these studies have conducted their training processes within the Gazebo simulation environment~\cite{gao2020deep, akmandor2022deep}. In~\cite{kaufmann2023champion}, a model-free, on-policy deep RL approach is employed to train the control policy within TensorFlow agents simulation. It aims to guide the drone at high speeds through gates while observing the current robot state's estimate, the gate's relative pose using an onboard camera, and the previous action.

A significant portion of existing studies, illustrated by the previous examples, are confined to simulations. Additionally, approaches are often tailored or the papers concentrate on either particular robots or specific considerations of a use-case. Furthermore, the lack of standardized benchmarks and open-source implementations hinders validation and comparison~\cite{kim2021reinforcement}. In this article, we address these gaps by providing detailed implementation steps, demonstrating the process of training an RL agent from simulation to real-world deployment on a generic wheeled robot. Specifically, through this magazine article, we also aim to introduce a broader audience to utilizing the state-of-the-art NVIDIA Omniverse Isaac Sim to achieve autonomous local planning and obstacle avoidance from the ground up. We delve into the key challenges and compare different approaches to training.
Through learning from robot's interactions with the environment—specifically by evaluating the presence and proximity of nearby walls and obstacles— these algorithms enable robots to make informed decisions and adapt to new situations, thereby improving their navigation capabilities in unknown environments. This makes reinforcement learning a potentially compelling approach for advancing mobile robot navigation.

The main contributions of this article with respect to the available literature are the following. First, we provide an in-depth description of the state-of-the-art Isaac Sim simulator for RL-based navigation of wheeled robots. Second, we discuss in detail different training strategies (e.g., curriculum learning) and the key aspects to account for when defining tasks, including reward function design. Finally, we demonstrate sim-to-real transfer of end-to-end RL policies across robots, while benchmarking to classical state-of-the-art navigation approaches. Overall, we aim to provide a more comprehensive analysis of the problem of training RL-based local planners for navigation than existing literature, making learning-based approaches to ground robot navigation accessible to a larger audience. Throughout the article, we assume familiarity with basic RL concepts.

The rest of the manuscript is organized as follows. Section~2 covers the use of Isaac Sim as a simulator for RL policy training. Section~3 then describes in more detail the different elements required to set up a training workflow. In Section~4, we delve into specific fine-tuning aspects of the proximal policy optimization (PPO) algorithm, one of the de-facto standards in the field, including reward modeling and model and training hyperparameters. Simulation and experimental results, with a focus on describing a sim-to-real transfer based on ROS\,2 and the Gazebo simulator, are introduced in Section 5. Section 6 discusses and concludes the work.

\section{RL With Isaac Sim}
\label{sec:design}
% Formerly, most RL robotics researchers heavily relied on CPU clusters to perform physically accurate simulations essential for training RL algorithms. However, after several years of NVIDIA’s research in GPU acceleration for RL, they have introduced the 
The introduction of the NVIDIA Omniverse Isaac Gym and Orbit frameworks have arguably aided in widening of audience and applications of deep reinforcement learning research. %These advancements make RL-based training for real-world robotics more accessible, replacing the need for dedicated hardware or software frameworks.

\subsection{Isaac Sim}

Isaac Sim, a GPU-based general-purpose physics simulation platform from Nvidia, serves as an extensible robotics simulator that empowers designers, researchers, and developers to create, test, and train AI-based robots such as wheeled robots, legged robots, and drones. Leveraging the power of NVIDIA Omniverse, Isaac Sim provides scalable, photorealistic, and physically accurate virtual environments for high-fidelity simulations. It can simulate realistic sensor models such as camera, lidar, and IMU, and a variety of objects and scenes, enabling tasks such as manipulation, navigation, synthetic data generation, and various computer vision applications through Python, ROS integration, and Isaac SDK.

\subsection{RL in Isaac Sim / Gym}

Omniverse Isaac Gym is an extension for reinforcement learning in robotics which is built on top of NVIDIA Isaac Sim. Isaac Gym is highly parallelized simulations by conducting both physics simulation and policy training on the GPU through an API, based on the vectorization of observations and actions. This framework offers a straightforward interface for training RL agents and supports various RL algorithms. In the latest releases of Isaac Gym, RL Games is introduced as the default library for running example environments. Whether it is training robotic agents to perform complex tasks, fine-tuning and optimizing RL policies, or evaluating their performance, Isaac Gym provides a bridge between the simulation environment and RL algorithms. 

More recent RL framework, also powered by NVIDIA Isaac Sim, are Orbit and Isaac Lab. Orbit offers a comprehensive suite of features, including support for various robot platforms, sensors, teleoperation, imitation Learning, and motion planning across diverse robotic applications, while Isaac Lab provides more comprehensive environments for different types of robots and tasks\footnote{\url{https://isaac-sim.github.io/IsaacLab/main/index.html}}. In these new frameworks, beginners may encounter a steeper learning curve. This article, therefore, focuses on the more documented and tested Isaac Gym as the first tutorial for straightforward path planning and navigation implementation, which can be further developed in Isaac Lab for more advanced features.

% \textbf{Other RL simulators}

% \red{EETU: mujoco? any other? Gazebo?}
% \newpage

\section{RL Workflow}

In Isaac Gym, the development and simulation of a customized RL agent navigation task needs a few basic steps for effective simulation, training, and testing. This includes defining the simulation environment and robot, crafting a Python script for the task class to specify goals, reward computation, and reset management, and utilizing two YAML configuration files—one for task parameters and another for training parameters—to complete the task. This section provides an overview of these essential core components. %s and explores their practical applications. % \hl{The implementation details are available in GitHub}\footnote{Code: \url{https://github.com/sahars93/RL-Navigation}}.
\subsection{Robot And Environments}

\begin{figure*}[t]
    \centering
    
    \begin{subfigure}[b]{.45\textwidth}
        \centering
        \includegraphics[height=5cm,width=0.9\textwidth]{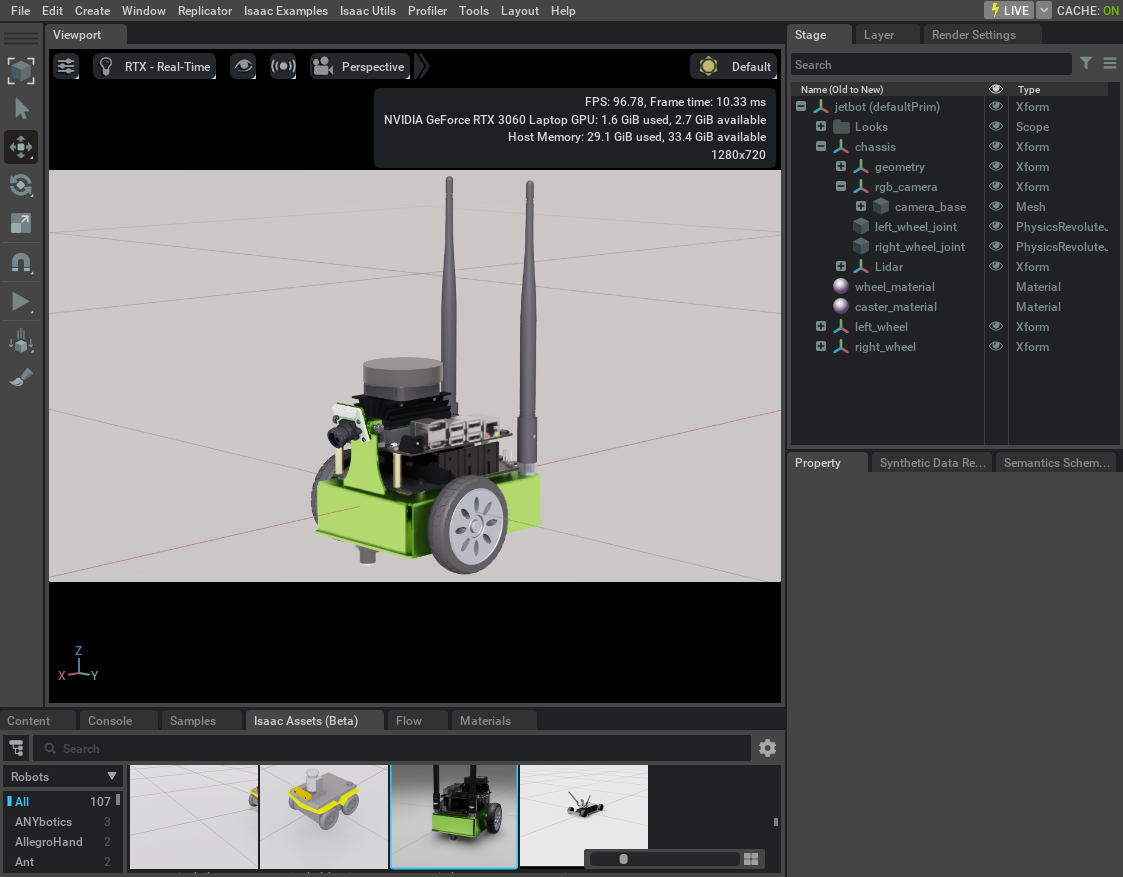}
        \caption{Isaac Jetbot robot}
        \label{fig:jetbot}
    \end{subfigure}
    \begin{subfigure}[b]{.45\textwidth}
        \centering
        \includegraphics[height=5cm,width=0.9\textwidth]{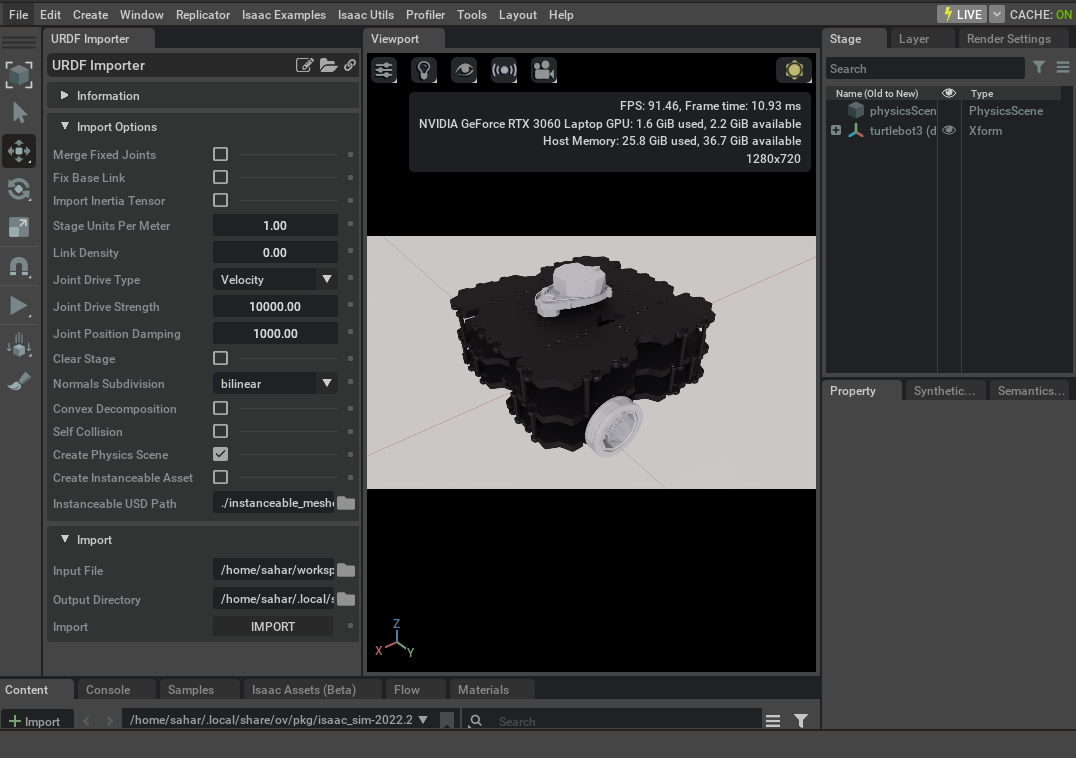}
        \caption{URDF Importer Extension}
        \label{fig:urdf2usd}
    \end{subfigure}
    
    \vspace{1em}
    \begin{subfigure}[b]{.45\textwidth}
        \centering
        \includegraphics[height=5cm,width=0.9\textwidth]{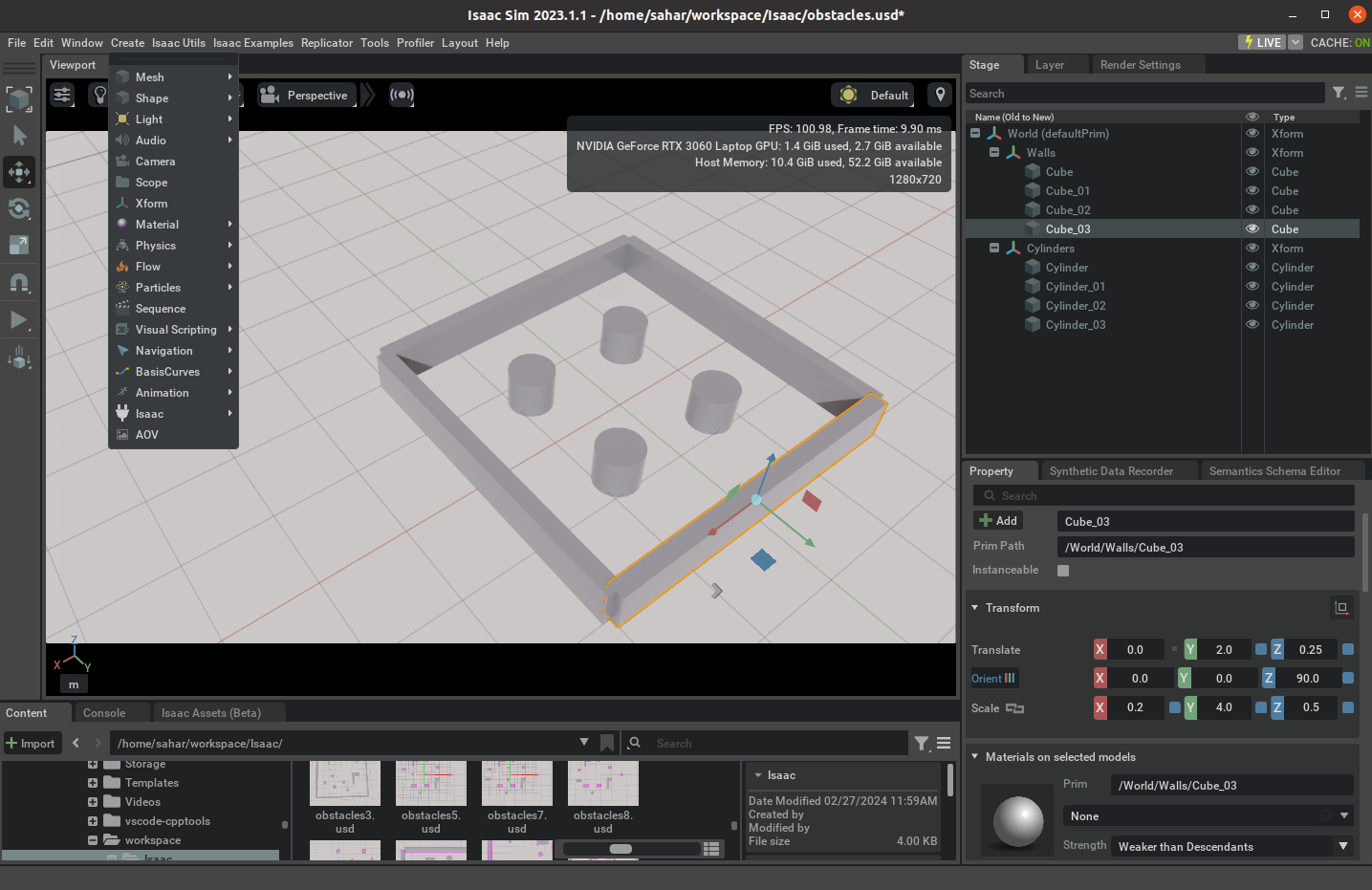}
        \caption{A custom environment in Isaac Sim}
        \label{fig:usd_scene}
    \end{subfigure}
    \begin{subfigure}[b]{.45\textwidth}
        \centering
        \includegraphics[height=5cm,width=0.9\textwidth]{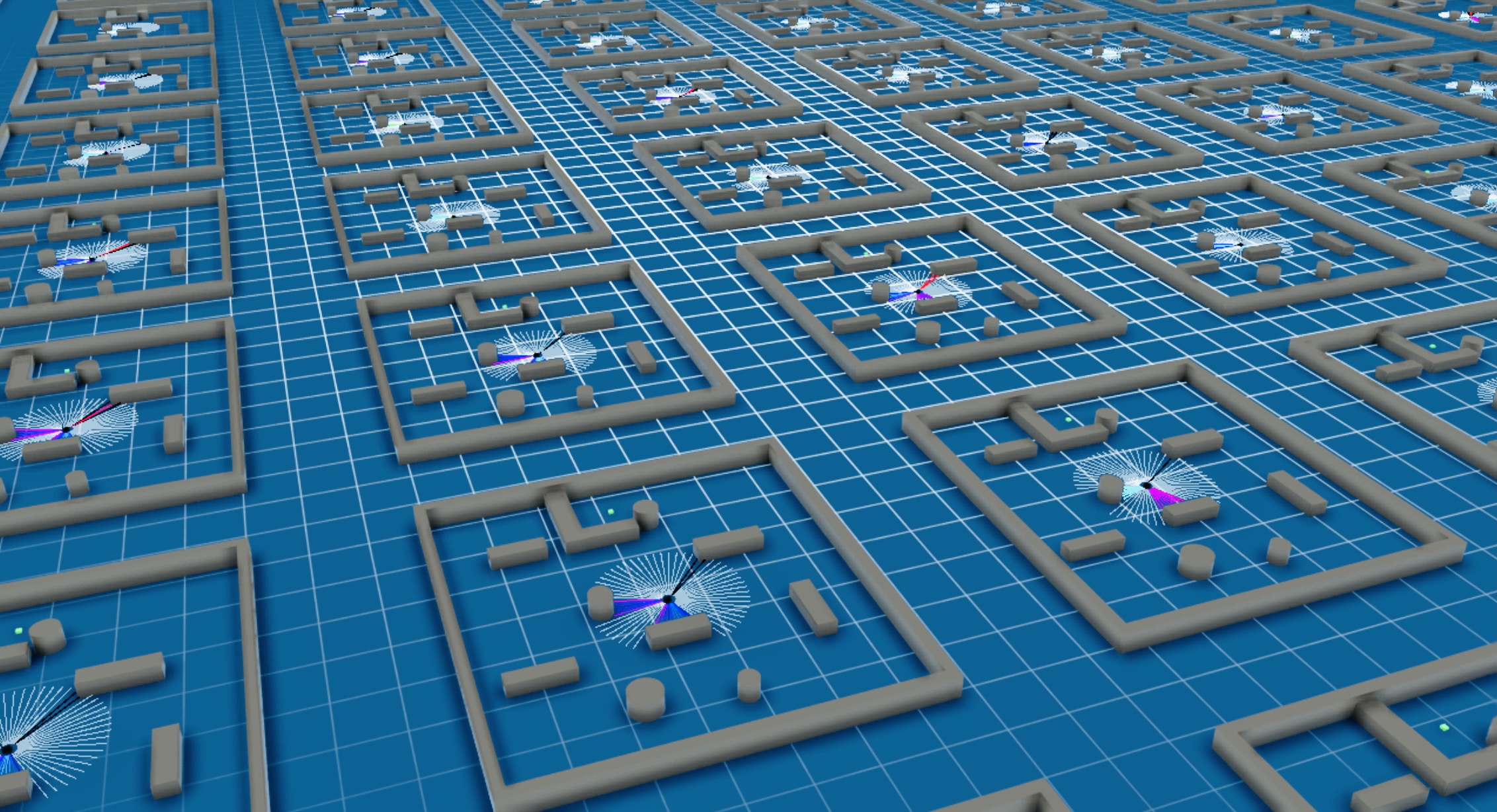}
        \caption{Isaac Gym}
        \label{fig:isaac_gym_env}
    \end{subfigure}
    \caption{Robots and Environments}
    \label{fig:robots}
\end{figure*}

Isaac Gym offers a user-friendly API for creating and configuring scenes with custom robots and objects. Many ROS users utilize the Unified Robot Description Format (URDF), a popular format for describing the basic robot cell and geometry, and practical applications. Isaac Sim supports various file formats, including URDF, Multi-Joint dynamics with Contact (MJCF), and Universal Scene Description (USD). It is feasible to incorporate custom robots from URDF and MJCF files into tasks, or alternatively, convert them to USD format using the Isaac Sim Importer extensions. Two robot models were employed in our mobile robot navigation task, the modified Isaac built-in Jetbot robot with Lidar sensor (Figure.\ref{fig:jetbot}) and the USD model of the Turtlebot3-Waffle from the TurtleBot3 Simulation ROS Package, converted through the Isaac URDF Importer as shown in Figure.~\ref{fig:urdf2usd} for additional experiments. To correctly import the mobile robot from the URDF file, the "Fixed Base Link" must be unchecked, and the "Joint Drive Type" must be set to "Velocity". When working on specific tasks, one can create, modify, and save custom scenes within Isaac Sim and add them as USD files into the task as shown in Figure.~\ref{fig:usd_scene}.

\subsection{Task}
A variety of reinforcement learning tasks are provided at Omniverse Isaac Gym extension, where main functionalities such as performing episode resets, applying actions, collecting observations, and computing rewards are implemented in this task class. Our $JetbotTask$, inherits from the BaseTask class in $omni.isaac.core$, comprises several key components. The general structure of the definition of each component of a new RL task is shown in Listing~\ref{lst:new_rl} The Initialization phase, the $init$ function, sets initial configurations for the environment. The initial setup for each task is detailed in its dedicated task YAML file. An overview of this file, along with a few sample parameters, is presented in~\ref{lst:task_yaml}. One can specify parameters related to the environment and simulation within this file. These include the number of environments, various sensors and USD configurations for the robot and objects, specifying CPU or GPU pipeline, and applying noise to observations, actions, and other properties in the domain randomization part. %These variables can also be specified directly as the command line arguments. 
Action and observation space and other parameters such as the episode's length are also defined in the init function. The $set\_up\_scene$ function is for setting up the scene by creating ArticulationView or RigidPrimView objects. This function involves defining the scene, sensor, and robot, or loading assets from USD, URDF, and MJCF file formats. The $get\_observation$ function generates the observation space using Lidar ranges, information on the target's relative position, and the robot's state. Computations required before stepping into the physics simulation, such as applying actions to move the robot based on policy decisions or resetting the environment, occur in the $pre\_physics\_step$ function. The calculation of rewards, resets, and extra buffers is handled in the $calculate\_metrices$ function. Finally, determining which environments need resetting is done in the last function of the task.
Besides the task config file, each task is accompanied by its configuration file containing training parameters such as the model and network structures, and the PPO parameters such as the learning rate, as shown in~\ref{lst:train_yaml}. These parameters are passed through $rlgames\_train.py$.

\definecolor{LightGray}{gray}{0.9}
\renewcommand{\figurename}{Code Snippet}

\begin{listing}
\caption{General structure of a new RL task definition in OmniIsaacGym.}
\includegraphics[width=.49\textwidth]{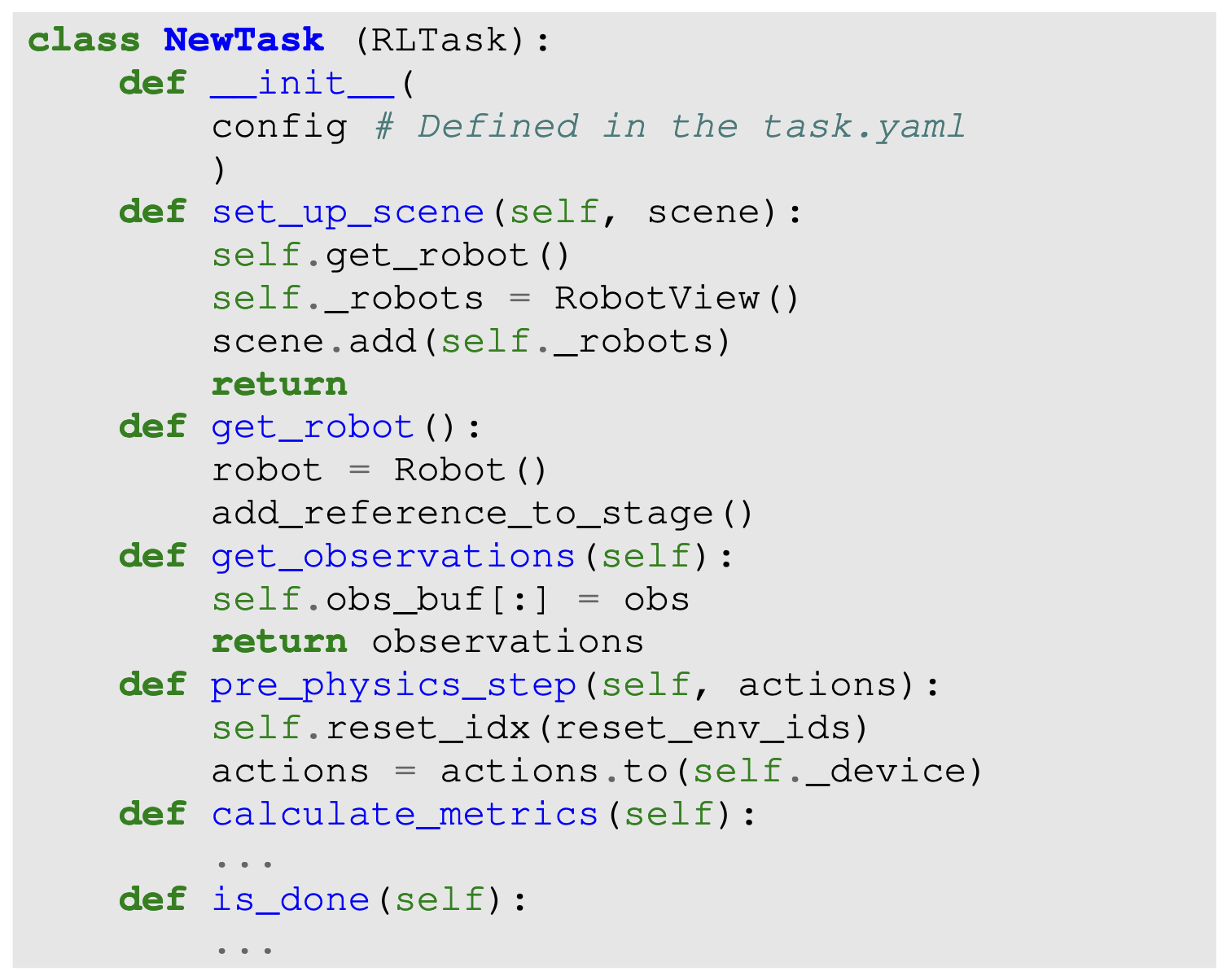}
\label{lst:new_rl}
\end{listing}

\renewcommand{\figurename}{Fig.}
\definecolor{LightGray}{gray}{0.9}
% \vspace{-1em}
\label{lst:task_yaml}
\begin{listing}
\caption{Core components of \textit{task.yaml}.}
\includegraphics[width=.48\textwidth]{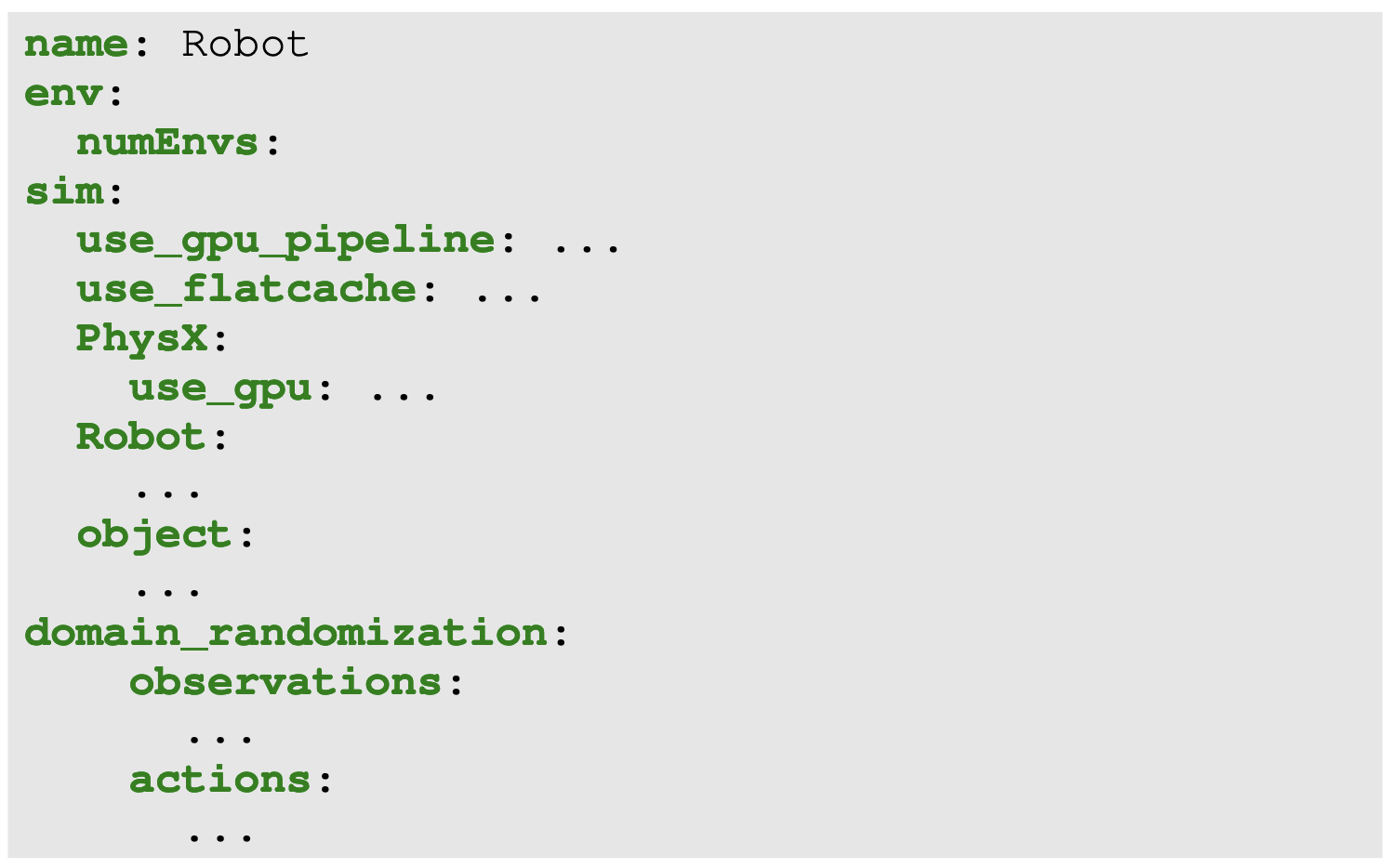}
\end{listing}
\definecolor{LightGray}{gray}{0.9}

\begin{listing}
\caption{Core components of train.yaml.}
\includegraphics[width=.48\textwidth]{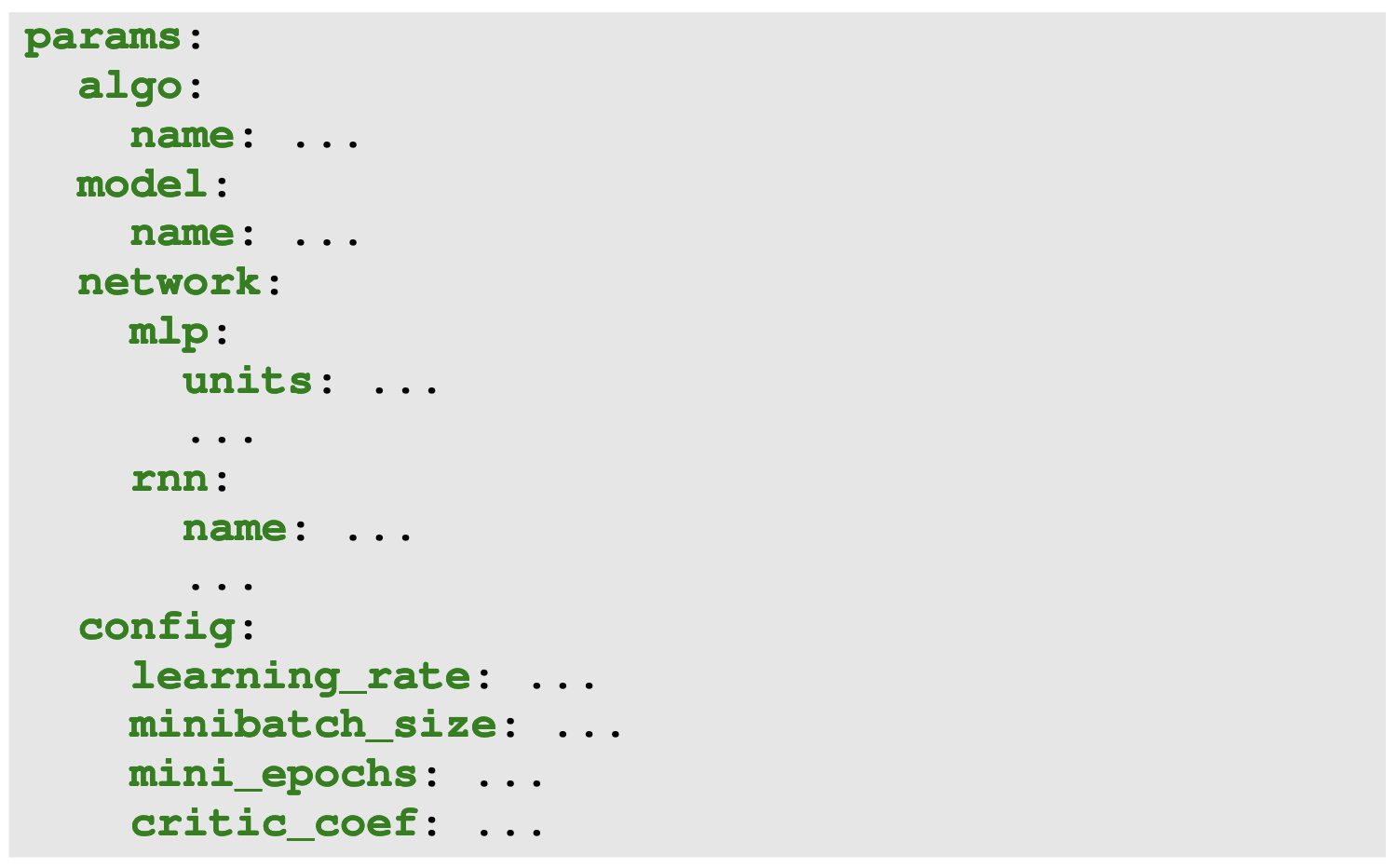}
\label{lst:train_yaml}
\end{listing}
\subsection{From Isaac To Gazebo}

% Robotic simulators offer a cost-effective and scalable alternative to the expensive process of testing, particularly in the context of RL tasks.
Before deploying navigation policies in the real world, we evaluate the trained policy within the Gazebo simulation environment. Gazebo, a widely adopted simulator in the global robotics community, directly interfaces with the ROS through user-friendly packages. This integration facilitates the creation of accurate simulations, and the outcomes obtained can be directly implemented on the real robots with only ROS installed, regardless of their software architecture. There are many Gazebo simulation packages available, featuring different differential drive robots equipped with Lidar sensors.  

In order to test in the Gazebo simulator, the trained model weights are exported in the format of ONNX model and integrated into a ROS node. The ONNX model can be imported and utilized for both simulation and real-world scenarios.

\subsection{ROS\,2 Node Deployment}

A ROS2 node is defined to handle the ROS operations within the robots. Upon initialization, such as the target's position, it sets up subscriptions to topics such as the Scan topic to get the Lidar ranges and the robot's position through the Odometry topic, along with a publisher for $cmd\_vel$ commands. Additionally, the class integrates an ONNX model, for inference. Various callback functions are implemented, such as $Odometry\_callback()$, and $scan\_callback()$, to process incoming messages from subscribed topics. Finally, the $send\_control()$ function generates robot control commands based on the model's outputs and publishes them via $cmd\_vel$.
\definecolor{LightGray}{gray}{0.9}

\begin{listing}[t]
\caption{ROS2 Node for real-time inference.}
\includegraphics[width=.48\textwidth]{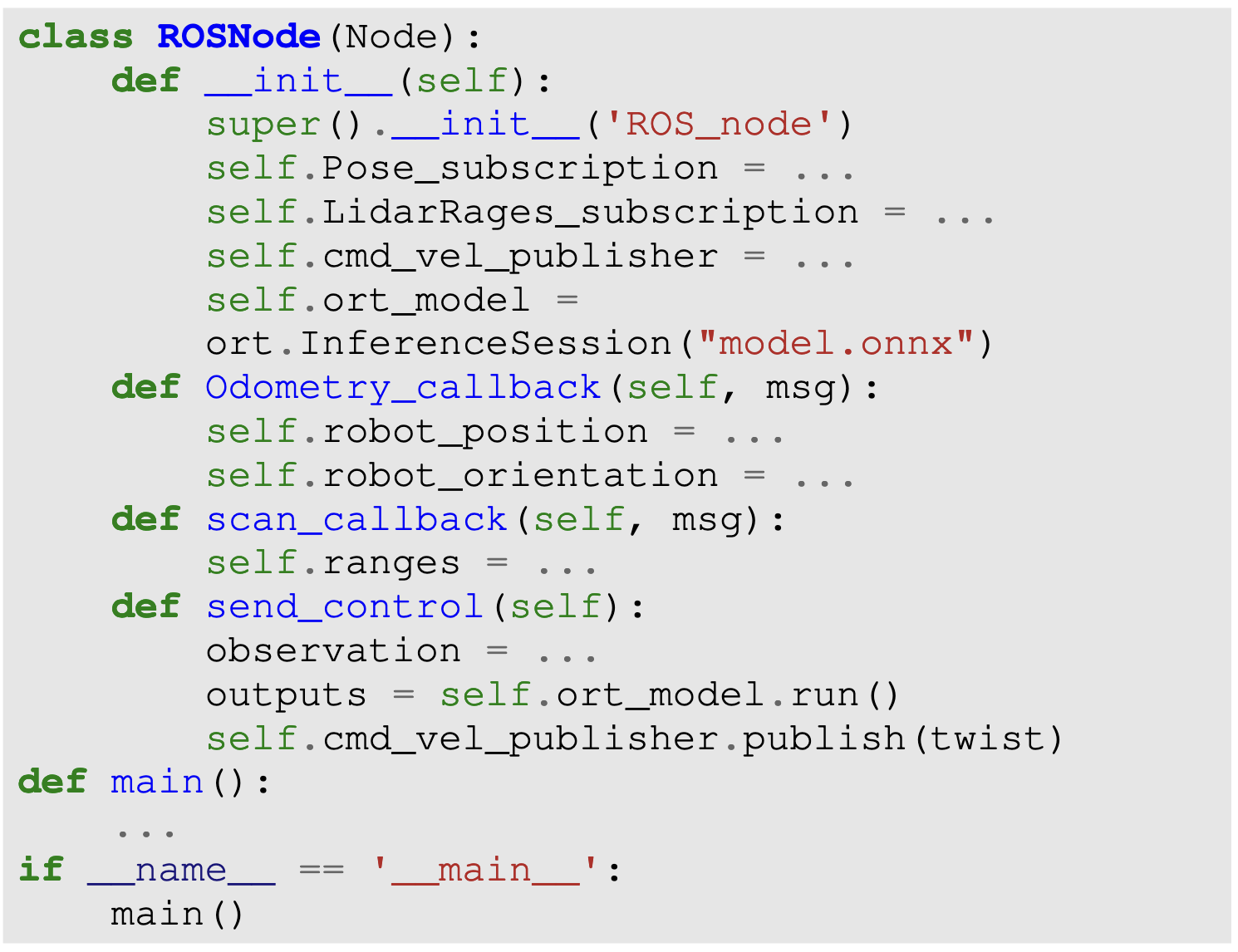}
\label{lst:ros_node}
\end{listing}

% \newpage

\section{RL for Navigation}%: Key Considerations}
\label{sec:approaches}

This section covers key aspects that need to be customized based on the robot, available sensors, or the overall use-case or general optimization objectives. These include the definition of observation and action spaces, reward modeling, and hyperparameter settings for both the model and the training and environment setup.

\subsection{Observation State}

The primary objective of an RL policy is to optimize the cumulative reward, by effectively navigating the interactions between the agent and its environment. In our navigation task, at each time step, the observation state $o_t$ comprises 2D lidar scans with 6-degree resolution (120 scans) in the range of [0.15, 3](m) shown as $L_t$, relative goal position as a 2D vector representing the robot's relative distance, $d_t$, and angle,  $\theta_t$, to the goal in polar coordinates, and the linear velocity $v_{t-1}$ from the previous time step, along with the angular velocity $\omega_{t-1}$. In~\eqref{eq:observation}, $a_t$ is the 2D action space, defined as linear velocity $v_t$ and angular velocity $\omega_t$, and the DRL policy, $\pi$, maps the observed state $o_t$, into the updated action, as the linear velocity in the range of [0.1, 0.5](m/s) and angular velocity in the range of [-0.5, 0.5](rad), to direct the robot towards its goal while avoiding collisions.

\begin{equation}
    \label{eq:observation}
    \begin{split}
    a_t &: {v_t, \omega_t} \\
    o_t &: {L_t, d_t, \theta_t, v_{t-1}, \omega_{t-1}}\\
     a_t &\sim \pi(o_t|a_t) 
\end{split}
\end{equation}

\subsection{Rewards}

In the reinforcement learning framework, the reward function plays a crucial role in assessing the effectiveness of the robot's actions. In navigation tasks, agents receive rewards based on goal achievement, collision avoidance, and time. Sparse rewards can hinder convergence, so the reward structure can be adjusted to better suit the task and enhance learning efficiency. In our navigation task, the primary aim is to ensure the robot avoids collisions and reaches its destination swiftly. The reward function, denoted as $R_t$ in~\eqref{eq:reward}, is designed with three main components. The robot receives rewards as it progressively reduces its distance from the target, defined as $r_{distance}$, and to avoid the collision, an exponential penalty, $r_{collision}$, is applied to the robot as it gets closer to the obstacles, increasing as it approaches the threshold $Min$, as the closest possible distance. Furthermore, to determine the shortest path, the robot is rewarded based on the time it takes to reach the target. This means that after reaching the target, the reward is calculated based on the remaining episode length.

\begin{equation}
\label{eq:reward}
R_t = \begin{cases}
r_{distance} &: +(d_{t-1} - d_t) \quad \\
r_{collision} &: -(e^{-{{min_{range}}}}) \quad \scriptstyle{{min_{range}} < Min} \\
r_{time} &: + \scriptstyle{({remaining \ steps})}\\
Goal &: +r_1 \quad \quad \scriptstyle{Reset: True}\\
Collision &: -r_2 \quad \quad \scriptstyle{Reset: True} \\
Max\ length &: -r_3 \quad \quad \scriptstyle{Reset: True}
\end{cases}
\end{equation}

In the reward function, we introduce additional conditions that signal the end of an episode. In addition to the previously discussed rewards, we define three fixed rewards that serve as reset flags. The first is awarded upon reaching the target, defined as $Goal$, $Collision$ is a fixed negative reward for collisions when the robot gets too close to obstacles beyond the $Min$ threshold, and $Max\ length$ is a penalty for surpassing the maximum episode length before reaching to the target. These three conditions collectively signify the conclusion of an episode in the $is\_done$ function in the Listing~\ref{lst:new_rl}. The reward amounts can be defined based on the size and type of the robot, avoidance distance, and the environment. In our Jetbot navigation task, the $Min$ is $25\ cm$ and all fixed rewards $r_1$, $r_2$, and $r_3$ are set to $30$.

In the literature, the three main components—goal, collision, and time—are essential, with their weights varying. A large penalty for collisions may limit exploration, while a high reward for the target might lead to overfitting to a specific target point, reducing generalization. Typically, training begins with equal weights, and adjustments are made based on the specific task.

\begin{figure}[t]
    \centering
    \setlength{\figurewidth}{0.48\textwidth}
    \setlength{\figureheight}{0.42\textwidth}
    \footnotesize{\input{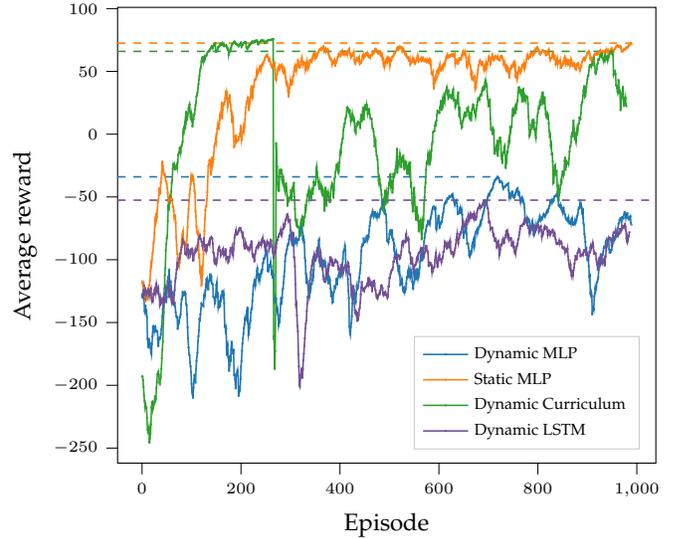}}
    \caption{Episodic returns during training using Isaac Sim}
    \label{fig:reward}
\end{figure}

\begin{figure}%{\subfigwidth}
    \centering
    \includegraphics[width=0.48\textwidth]{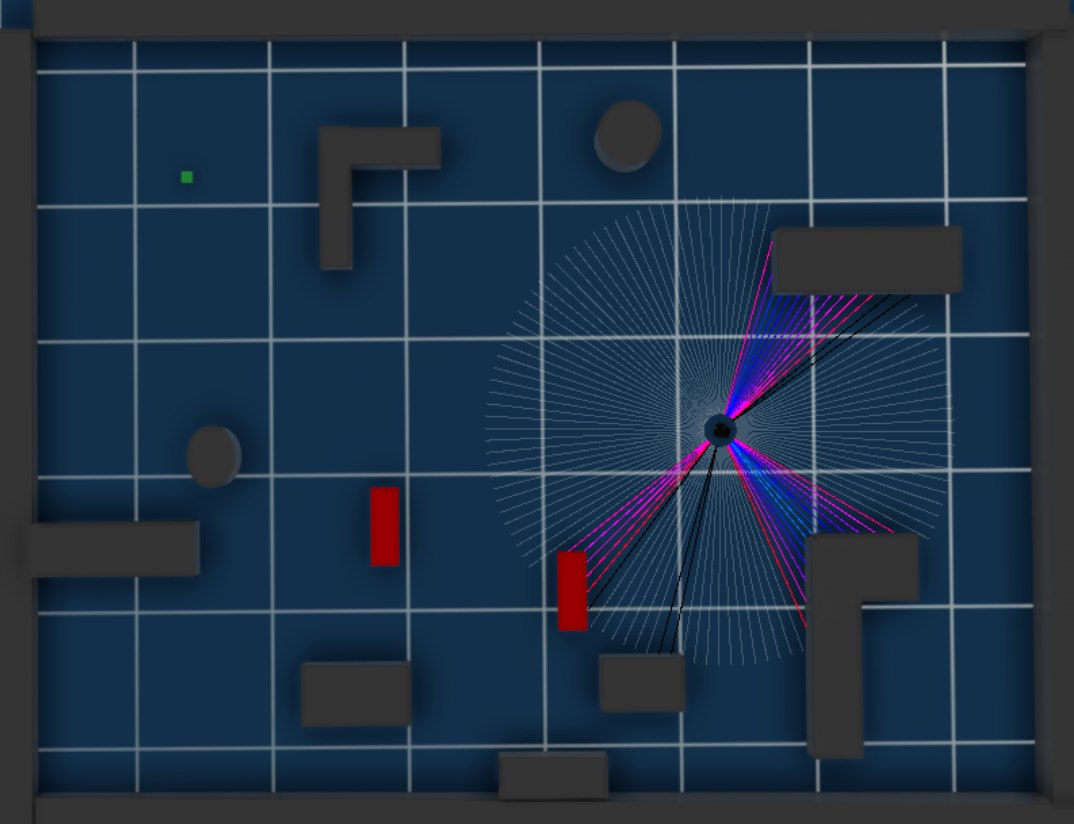}
    \caption{Illustration of the robot and lidar scan in an Isaac environment with dynamic obstacles pictured in red.}
    \label{fig:dynamic_env}
\end{figure}

\subsection{Model Definition}

We set the episode length to a maximum of 1200 steps, with 64 environments running for 1500 iterations. Our implementation revolves around the Actor-Critic algorithm designed for continuous spaces. The actor-critic algorithm is a reinforcement learning technique that merges policy-driven (Actor) and value-driven (Critic) approaches. The Actor selects actions per its policy, while the Critic assesses the Actor’s decisions. The Actor-Critic model utilizes a logarithmic standard deviation (logstd) for continuous action space, resulting in actions defined by a mean value with a fixed standard deviation. The network structure consists of a MultiLayer Perception (MLP) with 3 hidden layers of sizes $[256, 128, 64]$. Additionally, for training in the dynamic environment, we experimented with a Long Short-Term Memory (LSTM) layer of 128 hidden units after the input, followed by the MLP. As described earlier, these parameters are configured in the task and train config files. We train and test with two mobile robots, the Isaac built-in Jetbot, Figure~\ref{fig:jetbot}, and the Turtlebot3 robot imported from its URDF format, Figure~\ref{fig:urdf2usd}. When changing the robot, certain parameters in the initialization function must be adjusted, such as the wheel and speed settings in the $DifferentialController$ class or the minimum and maximum lidar scan ranges.

\section{Experimental Results}
\label{sec:results}

\subsection{Training In Isaac }
The training results, depicted in Figure~\ref{fig:reward}, show our experiments conducted in static and dynamic environments. In the static environment, the aforementioned MLP structure achieved a maximum of 73 rewards across 1000 episodes, signifying the successful accumulation of rewards for reaching the target in the shortest possible time. We employed the same network structure but added dynamic obstacles moving within the environment, red cubes in~Fig\ref{fig:dynamic_env}, which inherently complicated the training process. To further capture previous observations, we extended our training to include an LSTM layer followed by the same MLP layers. The training process in the dynamic environment shows limited promise, even when utilizing an LSTM layer. In fact, the robot fails to avoid obstacles and reach the target within 1000 episodes.

We adopted a curriculum learning approach for the dynamic environment to facilitate and accelerate the learning process, and enhance convergence. Curriculum learning is a training strategy that gradually increases the complexity of tasks or training samples. When performing an advanced navigation task, the initial step involves moving toward the target with associated rewards. Then, the environment’s complexity can be enhanced by including static obstacles, and a penalty for collisions in the reward function. Finally, dynamic obstacles are added during the training process. In our case, we initiated training with the simpler task, the static environment. This allowed the agent to learn initial policies for moving towards the target, resulting in faster convergence. After 300 initial steps, we added dynamic objects and continued the training process in a more complex environment. The staged curriculum approach enabled smoother adaptation to the dynamic obstacles, ultimately enhancing the agent’s performance and episode return.

\begin{figure}[t]
    % \centering
    \setlength{\figurewidth}{0.48\textwidth}
    \setlength{\figureheight}{0.6\textwidth}
    \footnotesize{\input{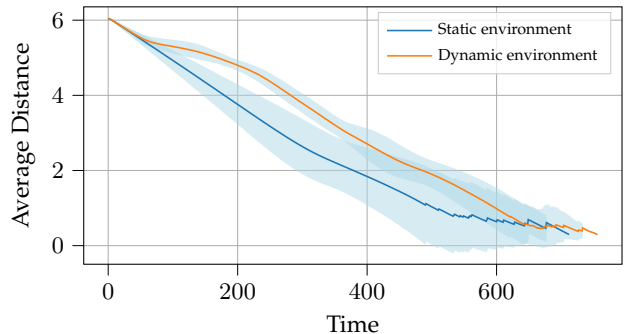}}
    \caption{Average and standard deviation of distance to the target as a function of time during tests using Jetbot in Isaac Sim with static and dynamic environments.}
    \label{fig:distance_time}
\end{figure}

\begin{figure}[t]
    \centering
    
    % Define the width and height for all subfigures
    \newlength{\subfigwidth}
    \setlength{\subfigwidth}{0.42\textwidth}
    % \newlength{\subfigheight}
    % \setlength{\subfigheight}{0.3\textwidth}
    
    % \begin{subfigure}[b]{\subfigwidth}
    %     \centering
    %     \includegraphics[%height=\subfigheight,
    %     width=\textwidth]{fig/dynamic.png}
    %     % \captionsetup{skip=50pt}
    %     \caption{Dynamic obstacles}
    %     \label{fig:dynamic_env}
    % \end{subfigure}
    % \hfill % This will add space between the subfigures if needed
    
    \begin{subfigure}[b]{\subfigwidth}
        % \centering
        \setlength{\figurewidth}{.91\textwidth}
        \setlength{\figureheight}{.91\textwidth}
        \footnotesize{\input{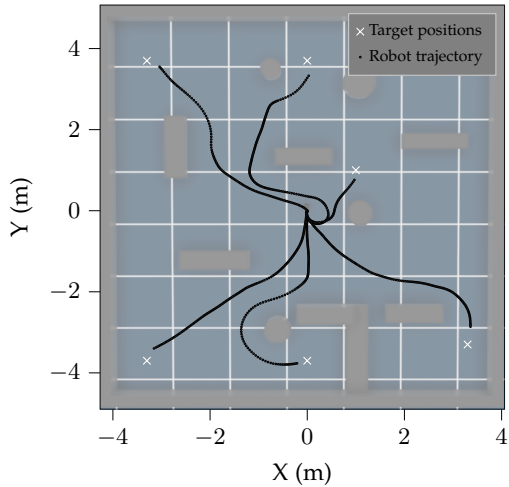}}
        \vspace{-.6em}
        \caption{Jetbot trajectory, different targets (static environment)}
        \label{fig:jetbot_traj}
    \end{subfigure}

    \vspace{.45em}
    
    % \vspace{1em} % This adds vertical space between the rows of subfigures
    \begin{subfigure}[b]{\subfigwidth}
        \centering
        \setlength{\figurewidth}{.91\textwidth}
        \setlength{\figureheight}{.91\textwidth}
        \footnotesize{\input{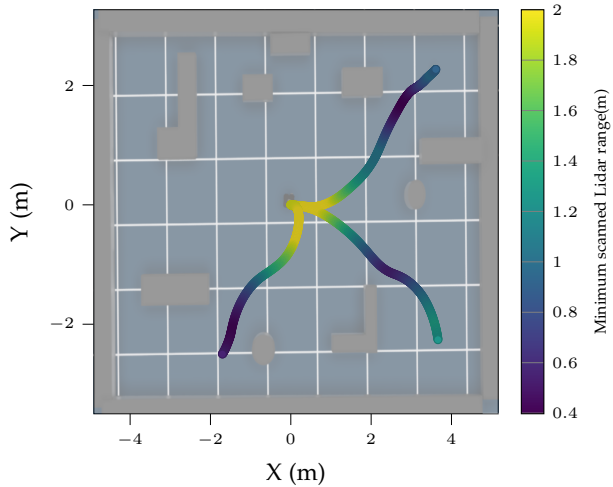}}
        \vspace{-1em}
        \caption{Turtlebot3 trajectory, safety spectrum (static environment).}
        \label{fig:turtle_traj}
    \end{subfigure}

    \vspace{.45em}
    % \hfill
    
    \begin{subfigure}[b]{\subfigwidth}
        \centering
        \setlength{\figurewidth}{.91\textwidth}
        \setlength{\figureheight}{.91\textwidth}
        \footnotesize{\input{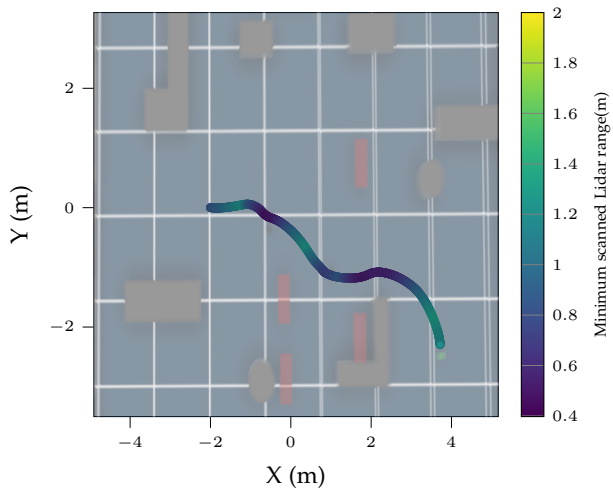}}
        \vspace{-1em}
        \caption{Jetbot trajectory in dynamic environment.}
        \label{fig:dynamic_traj}
    \end{subfigure}
    
    \caption{Qualitative evaluation performance with minimum lidar range throughout the trajectories in Isaac Sim.}
    \label{fig:IsaacTest}
\end{figure}

% \begin{figure}[t]
%     \centering
%     \includegraphics[width=0.4\textwidth]{fig/dynamic.png}
%     \caption{Dynamic obstacles}
%     \label{fig:dynamic_env}
% \end{figure}

% \begin{figure}[t]
%     \centering
%     \setlength{\figurewidth}{0.48\textwidth}
%     \setlength{\figureheight}{0.36\textwidth}
%     \input{tex/isaac_trajectory}
%     \caption{Jetbot trajectory toward different targets}
%     \label{fig:jetbot_traj}
% \end{figure}

% \begin{figure}[t]
%     \centering
%     \setlength{\figurewidth}{0.48\textwidth}
%     \setlength{\figureheight}{0.36\textwidth}
%     \input{tex/turtlebot_trajectory}
%     \caption{Turtlebot3 Trajectory with safety spectrum }
%     \label{fig:turtle_traj}
% \end{figure}

% \begin{figure}[t]
%     \centering
%     \setlength{\figurewidth}{0.48\textwidth}
%     \setlength{\figureheight}{0.36\textwidth}
%     \input{tex/dynamic_trajectory}
%     \caption{Robot Trajectory in dynamic environment with safety spectrum }
%     \label{fig:dynamic_traj}
% \end{figure}

\subsection{Isaac Simulations}

Within the process of validating our model, a series of tests were conducted within the Isaac simulation environment, utilizing both Jetbot and Turtlebot3 robots. These tests involved setting various target positions to assess the model’s adaptability. Figure~\ref{fig:jetbot_traj} illustrates the Jetbot navigating through a novel environment, adeptly heading towards diverse target poses while avoiding obstacles along multiple paths. For Turtlebot3, Figure~\ref{fig:turtle_traj} conveys the trajectory outcomes, where the proximity to obstacles is indicated by a gradient of colors, reflecting the robot’s distance to the nearest object. This visual representation underscores the model’s proficiency in maintaining a safe distance from obstacles, a critical aspect of autonomous navigation. Extending the assessment to dynamic settings, Figure~\ref{fig:dynamic_traj} depicts robot’s successful navigation past moving objects, effectively avoiding collisions. This performance is attributed to the model’s curriculum-based training, as discussed in the preceding section. 

% \begin{figure*}[t]
%     \centering
%     \setlength{\figurewidth}{0.48\textwidth}
%     \setlength{\figureheight}{0.36\textwidth}
%     \input{fig/3models.png}
%     \caption{Comparative performance analysis of the RL policy and Nav2 Stack in Gazebo.}
%     \label{fig:distance_time_gazebo}
% \end{figure*}

\begin{figure*}[t]
    \centering
    \includegraphics[width=.95\textwidth]{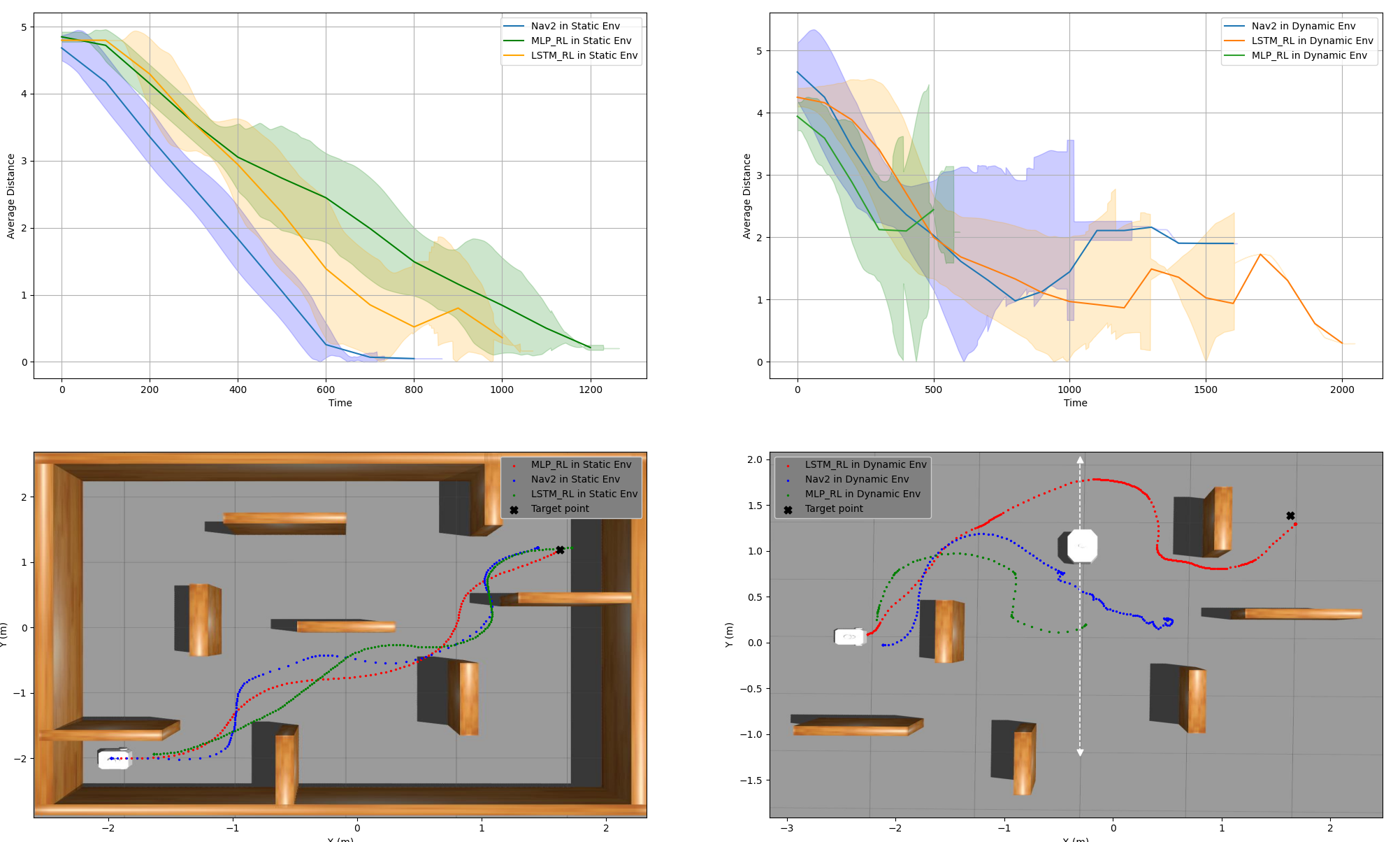}
    \caption{Comparative performance analysis of the RL policy and Nav2 Stack in Gazebo.}
    \label{fig:distance_time_gazebo}
\end{figure*}

Finally, Figure~\ref{fig:distance_time} presents an aggregation of the robots’ relative distances to a specified target from Figure~\ref{fig:dynamic_traj}, compiled over 30 trials. This data encompasses scenarios with and without dynamic obstacles, providing a comprehensive view of the model’s efficacy in variable environments.

% \begin{figure*}[t]
%     \centering
%     \setlength{\figurewidth}{0.98\textwidth}
%     \setlength{\figureheight}{0.36\textwidth}
%     \input{tex/gazebo_distance}
%     \caption{Distance to the target}
%     \label{fig:reward}
% \end{figure*}

\subsection{Gazebo Simulations}

Prior to real-world deployment, the RL model was converted to an ONNX format and subjected to tests in various Gazebo environments using a ROS\,2 node. We conducted a comparative analysis of the LSTM- and MLP-based RL models against Nav2~\cite{macenski2023desks}, the de facto ROS\,2 navigation stack, a sophisticated control system designed for autonomous robot navigation to a goal state based on the robot's current position, a map, and a target location. Figure~\ref{fig:distance_time_gazebo} illustrates the comparative results in both static and dynamic settings.

In the experiments, a mobile box was placed ahead of the robot, with its velocity adjusted between 0.1 and 0.3 m/s across various trials to obstruct the robot’s trajectory toward the designated target. Over 10 iterations of the same path, the distance-to-target distribution over time was comparable across all three methods (MLP-, LSTM-based RL, and Nav2) in static environments. This outcome is expected, as the Nav2 stack relies on a precomputed cost map, which can be less effective when encountering dynamic obstacles or sudden environmental changes, potentially leading to mission failure.

In contrast, the LSTM-based RL model, specifically trained to handle such scenarios, demonstrated superior performance. It consistently navigated around obstacles and avoided collisions, showcasing its robustness in dynamic conditions and its potential for reliable real-world applications. Meanwhile, the one-step training model using the MLP exhibited reduced performance in dynamic obstacle avoidance, underscoring its limited effectiveness for this task.

% \subsection{Testing in Gazebo }

% \begin{figure*}[t]
%     \centering
%     \includegraphics[width=0.95\textwidth]{fig/three_env.png}
%     \caption{Caption.}
%     \label{fig:diagram}
% \end{figure*}

% \begin{figure*}[t]
%     \centering
%     \includegraphics[width=0.95\textwidth]{fig/gazebo_test.png}
%     \caption{Caption.}
%     \label{fig:diagram}
% \end{figure*}

% \newpage
\subsection{Real-World Validation}

\begin{figure}[t]
    \centering
    \includegraphics[width=.32\textwidth]{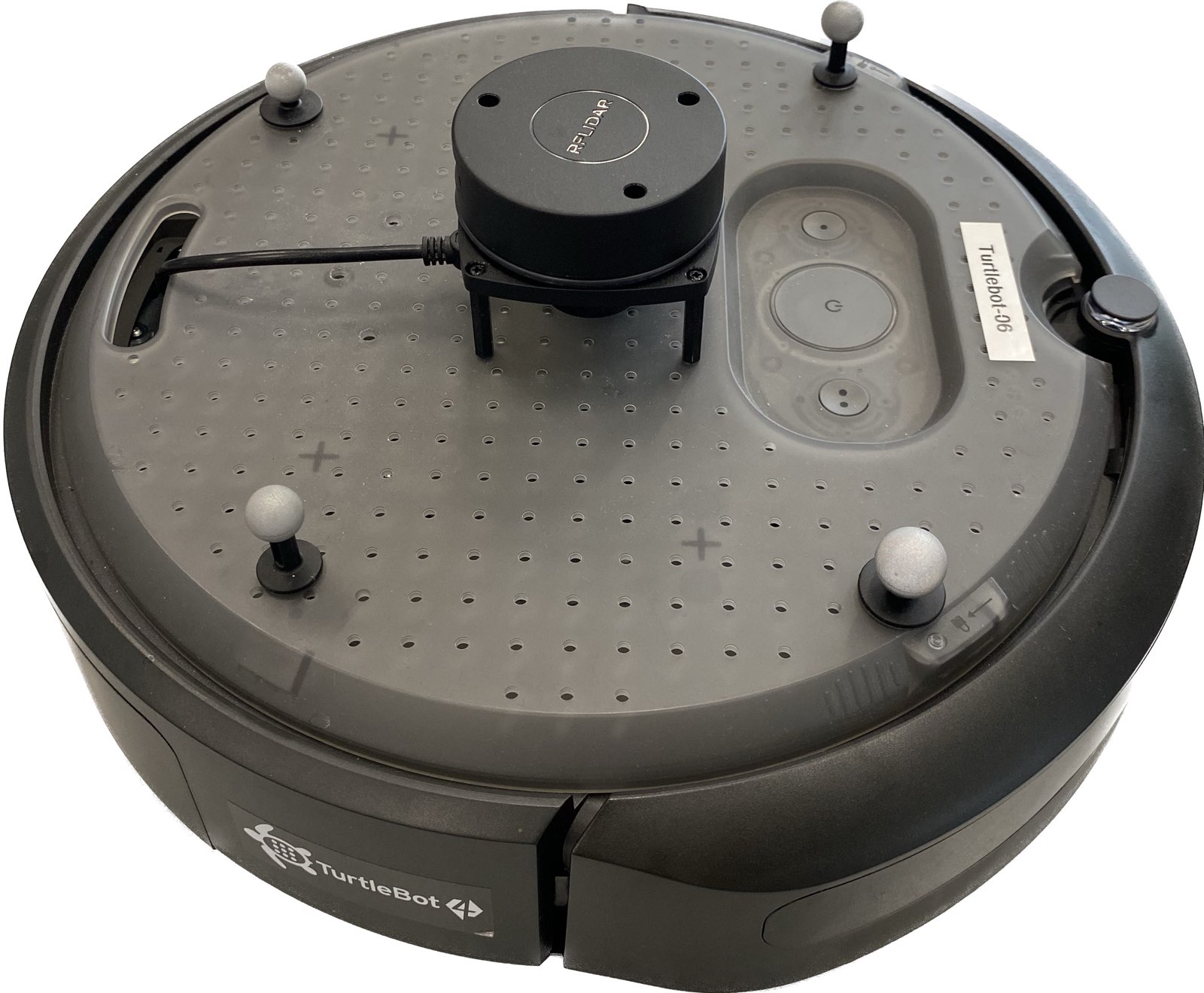}
    \caption{TurtleBot 4 Lite robot with 2D lidar used for the real-world experimental evaluation.}
    \label{fig:lite}
\end{figure}

\begin{figure*}[t]
    \centering
    \setlength{\figurewidth}{0.28\textwidth}
    \setlength{\figureheight}{0.28\textwidth}
    \footnotesize{\input{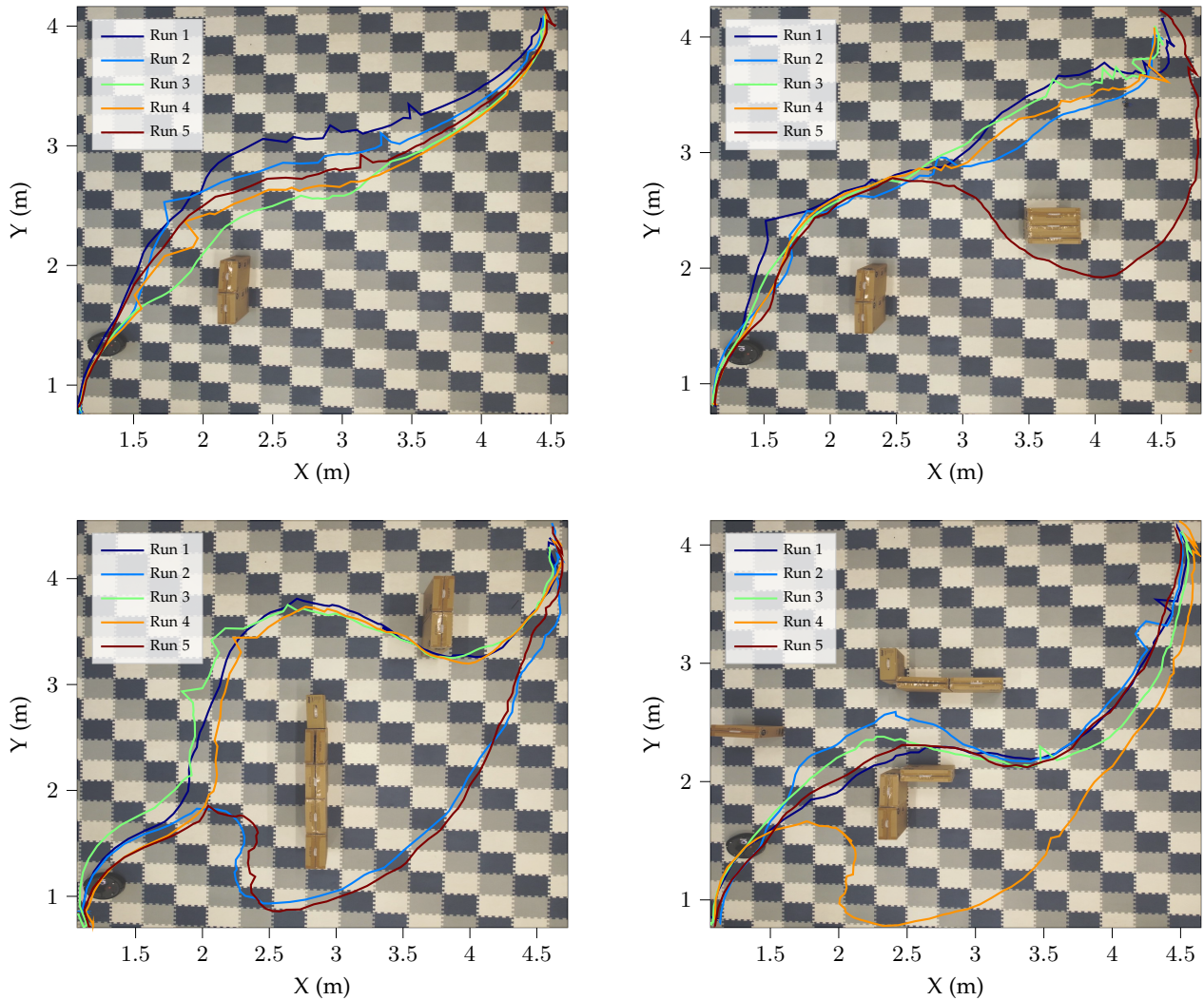}}
    \caption{Navigation in Various Real-World Environments}
    \label{fig:real4exp}
\end{figure*}

\begin{figure*}[t]
    \centering
    \includegraphics[width=0.8\textwidth]{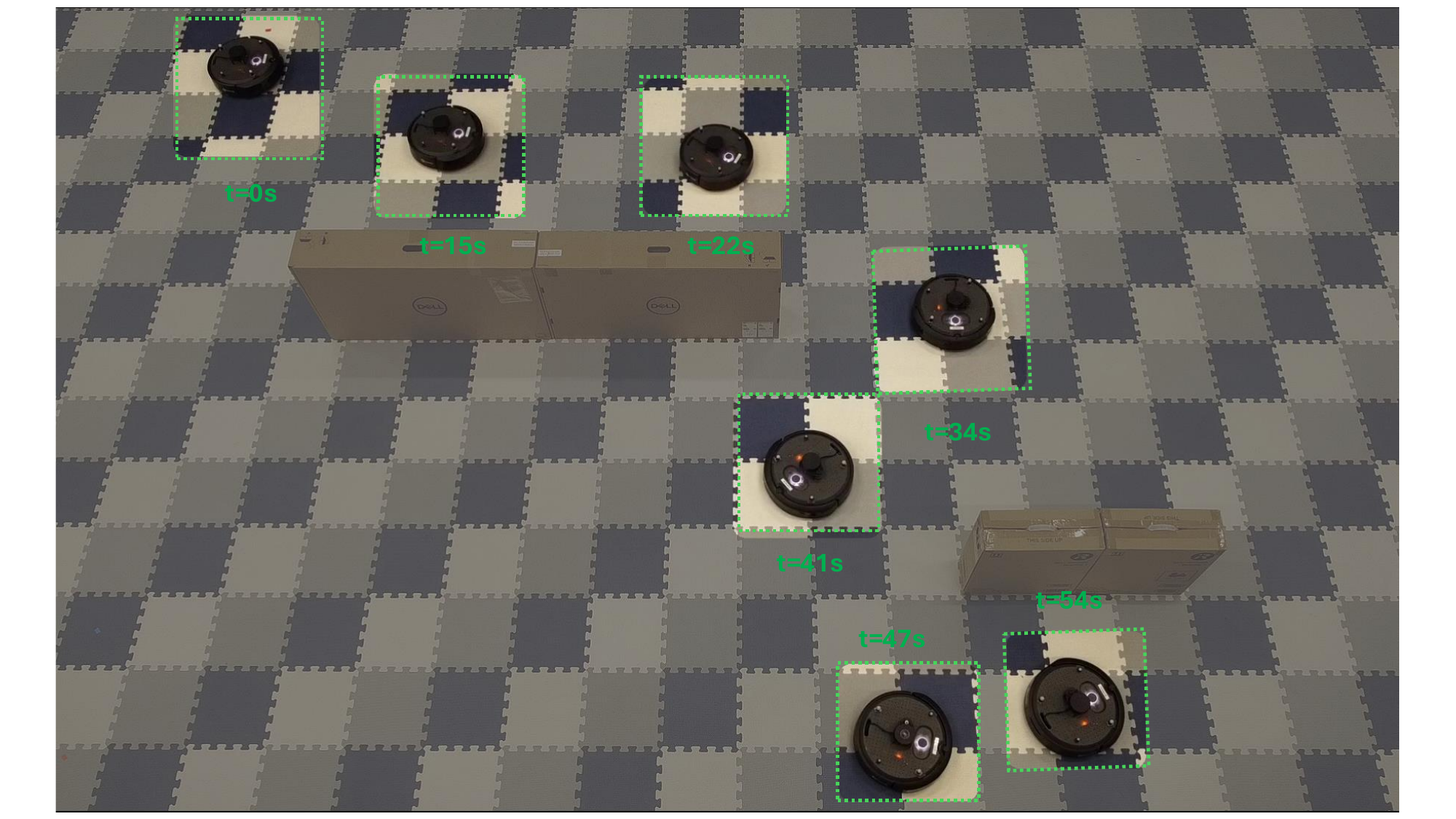}
    \caption{Robot trajectories with timestamps showing obstacle avoidance and navigating towards the target.}
    \label{fig:trajectory_timestamps}
\end{figure*}

\begin{table*}[t]
    \centering
    \caption{\small Real-world performance statistics over 10 trials for 4 different experiments.} 
    \label{table:performance}
    \normalsize{
    \begin{tabular}{@{}lcccccccc@{}}  
        \toprule
        & Task & Success & Min. lidar & Avg. linear & Dist. to   \\
        & time (s) & rate & range (m) & vel. (m/s) & target (m) \\
        \midrule   
        \textbf{Exp. 1}      & 39 & 10/10     & 0.43       & 0.15  & 4.69 \\ [+0.2em] 
        
        \textbf{Exp. 2}     & 41  & 10/10      & 0.39       & 0.15  & 4.70 \\ [+0.2em]  

        \textbf{Exp. 3}     & 53  & 8/10      & 0.40       & 0.13  & 5 \\ [+0.2em]  
        
        \textbf{Exp. 4}  &  52 & 7/10           & 0.25        & 0.14  & 4.73  \\
        \bottomrule
    \end{tabular}
    }
\end{table*}

\begin{figure*}[t]
    \centering
    \setlength{\figurewidth}{0.42\textwidth}
    \setlength{\figureheight}{0.3\textwidth}
    \footnotesize{\input{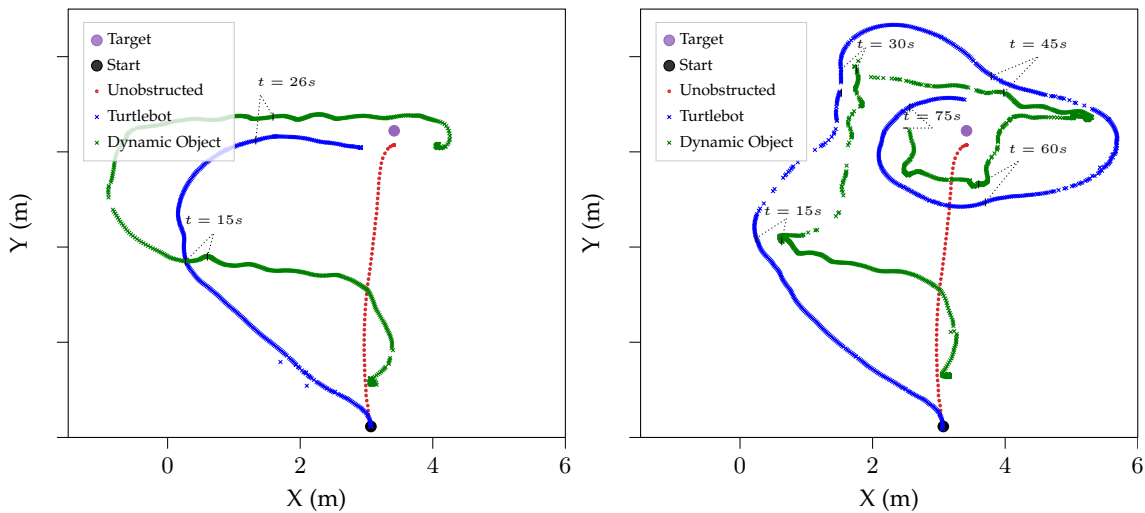}}
    \caption{Trajectories of Turtlebot in the presence of a dynamic obstacle in real-world experiments.}
    \label{fig:dynamic_real}
\end{figure*}

The approach adopted in this research is adaptable to various mobile robotic systems. For the practical implementation and assessment of the RL navigation model, we employed the TurtleBot 4 Lite, which was equipped with an RPLIDAR A1M8, granting a complete 360-degree perspective, and interfaced with a Raspberry Pi 4B as the On-board Computer, Figure~\ref{fig:lite}. The dimension of the TurtleBot 4 Lite is 342 x 339 x 192 mm, with wheels measuring 72 mm in diameter. Its maximum safe mode linear velocity is 0.31 m/s. The LiDAR has a minimum detection distance of 0.15 m, with its range configured up to 2 m and a resolution of 3 degrees.
% The hardware specifications of the robots, including a customized configuration of the RPLIDAR’s range, are detailed in Table~\ref{table:lite}.

The refined model weights have been exported and integrated into a ROS2 Galactic node. The control system runs on the Raspberry Pi On-board Computer. The robot’s positioning data was captured using an Optitrack Motion Capture system. To evaluate the system in real-world static environments, we designed test scenarios with obstacles of varying sizes and shapes, ensuring these differed from those used during training to assess generalization. Additionally, the robot's starting position was carefully chosen to align with the training setup, where it consistently began in relatively free space. This was necessary to avoid requiring additional training steps to account for diverse initial conditions. Figure~\ref{fig:real4exp} illustrates the robot’s navigation paths in four distinct settings with varied obstacles. As shown in Figure~\ref{fig:trajectory_timestamps}, the robot's trajectories at different timestamps illustrate its movement while navigating towards the target. We conducted 10 trials for each test, and the bolded average performance metrics are presented in Table~\ref{table:performance}. The results from the successful trials indicate that the robots maintained a minimum distance of 25 cm from obstacles, aligning with the training threshold. This safety margin can be adjusted to suit different robot dimensions and configurations. Similarly, the linear velocity parameter is flexible and can be tailored as required.

We expanded our experimentation to include dynamic obstacles—specifically, people—that obstruct the robot’s path toward the target. The robot's trajectories, both without obstacles and with dynamic obstacles, are illustrated in Figure~\ref{fig:dynamic_real}. To highlight the robot’s avoidance behavior, we marked the positions of the robot and obstacle every 15 seconds, showing when the robot adjusts its path. The individuals moved unpredictably in front of the robot, partially obstructing its direct path toward the target. At times, the person would suddenly appear in the robot’s trajectory, while in other instances, they moved alongside the robot, attempting to obstruct its path. This random movement pattern was designed to simulate real-world  scenarios with a priori unpredictable dynamic obstacles.

While curriculum learning improves performance in dynamic environments, models often struggle with tasks beyond their training conditions in dynamic environments. Handling diverse obstacle sizes, shapes, directions, and speeds requires carefully tuned reward functions and additional training stages to maintain robust performance.

% \section{Outline}

% \textbf{Questions:}
% Who are the beneficiaries?
% Robotics Researchers
% Clinical researchers

% \textbf{Research Questions:}
% What do we need to measure from humans to achieve HRI?
% How can we measure the human in the loop? 
% Contribution: ROS-health architecture and demo
% Models of human response to robot interactions.

% Methods: CONCEPT: ROS healthcare
% sharing information about human life through ROS
% ROS-Health structure and software architecture 
% Validation Experiments:
% DEMO:
% Sensors:
% ECG (VivaLink)
% PPG (Corsano)
% Pressure Mats x2
% Fitbit - Sleep Stages - 
% IMU
% Biosignals:
% HR
% RR
% HRV
% ECG x 12
% PPG x n
% Sleep
% ADLs
% Posture
% Robot 1: Wheelchair tracking: 
% Whill autonomous driving → Person state: ADLs + HR + Stress Level (EDA + HRV) 
% Robot 2.: Somnomat tracking:
% Person’s state during the night - sleep and HR

\section{Discussion}
\label{sec:discussion}

We tested the model in various static environments across simulators and the real world, achieving promising results. However, its generalization and robustness faced challenges in dynamic real-world settings. The presence of diverse conditions and noise—such as changes in the size, shape, speed, and direction of dynamic obstacles—highlighted the need for further adjustments. To improve performance in such scenarios, retraining, progressive training, and potential modifications to the reward function may be necessary. Additionally, real-world test feedback led to fine-tuning parameters, such as adjusting the LiDAR sampling process to account for narrower dynamic obstacles. These adjustments required either retraining from the baseline or more training to refine the reward function. To enable safe and effective navigation through dynamic obstacles such as humans, incorporating a specialized reward function term, such as a social-safety zone, may be essential for fostering human-aware navigation. This approach can be adapted and generalized for other specific environments requiring tailored modifications.

% \newpage
\section{Conclusion}
\label{sec:conclusion}

Throughout this article, we have described the process of setting up a training workflow for a RL policy for mobile robot navigation in Isaac Sim. We covered the key steps in defining the robot model, and the training environment and RL task. Additionally, we discuss important aspects in terms of hyperparameter tuning from the perspective of both the training setup and the actual policy model. Finally, we describe the workflow to enable the transfer from simulation to reality, with examples and ROS\,2 node templates.

To demonstrate the effectiveness and usability of such RL policies for real-world deployments, we also analyze the performance both quantitatively and qualitatively in simulation (Isaac Sim and Gazebo) and real-world experiments. We use Nav2, the de-facto standard ROS\,2 navigation stack, as a benchmark in a subset of the simulations.  

The experimental results convey that state-of-the-art performance can be obtained by training fully in the simulation environment. This opens the door to quick deployment of new robots with end-to-end RL-based control, where both perception, trajectory planning and tracking are encapsulated in a single process. While this does not necessarily offer the best fine-tuned performance, we believe this to be a step towards, e.g., low-code applications.

Overall, this work aims to discuss the different possible approaches to RL-based local navigation and obstacle avoidance, beyond the specific approaches and use-cases widely showcased in the literature. This has been presented in an instructional style, but also covering new experimental results. Importantly, we believe this work fills a gap in the literature in terms of the introduction of generic sim-to-real workflows and a generalizable approach to RL navigation in mobile robotics.

%%%%%%%%%%%%%%%%%%%%%%%%%%%%%%%%%%%%%%%%%%%%%%
%%                                          %%
%%            ACKNOWLEDGMENT                %%
%%                                          %%
%%%%%%%%%%%%%%%%%%%%%%%%%%%%%%%%%%%%%%%%%%%%%%

\section*{Acknowledgments}

This work was supported by the R3Swarms project funded by the Technology Innovation Institute (TII).%, and by the Innosuisse Project for “Developing an AI-enabled Robotic Personal Vehicle for Reduced Mobility Population in Complex Environments“ under agreement no. 2155012544.

%%%%%%%%%%%%%%%%%%%%%%%%%%%%%%%%%%%%%%%%%%%%%%
%%                                          %%
%%              BIBLIOGRAPHY                %%
%%                                          %%
%%%%%%%%%%%%%%%%%%%%%%%%%%%%%%%%%%%%%%%%%%%%%%
% \newpage
% \nocite{*}
% \red{Limit of 20 references}
\bibliographystyle{ieeetr}
\bibliography{bibliography}

\end{document}